\documentclass[a4paper,twoside]{article}
\usepackage{soul}
\usepackage{epsfig}
\usepackage{makecell}
\usepackage{subcaption}
\usepackage{calc}
\usepackage{amssymb}
\usepackage{amstext}
\usepackage{amsmath}
\usepackage{amsthm}
\usepackage{multicol}
\usepackage{pslatex}
\usepackage{apalike}
\usepackage{algorithm2e}
\usepackage[bottom]{footmisc}
\usepackage{SCITEPRESS}     
\begin{document}

\title{Real-Time Bus Arrival Prediction: A Deep Learning Approach for Enhanced Urban Mobility}

\author{\authorname{Narges Rashvand\sup{1}, Sanaz Sadat Hosseini\sup{2}, Mona Azarbayjani\sup{3}, and Hamed Tabkhi\sup{1}}
\affiliation{\sup{1}Department of Electrical and Computer Engineering, University of North Carolina at Charlotte, Charlotte, NC, USA}
\affiliation{\sup{2}Department of Civil and Environmental Engineering, University of North Carolina at Charlotte, Charlotte, NC, USA}
\affiliation{\sup{3}Department of Architecture, University of North Carolina at Charlotte, Charlotte, NC, USA}
\email{\{nrashvan, shossei7, mazarbay, htabkhiv\}@charlotte.edu}}


\abstract{In urban settings, bus transit stands as a significant mode of public transportation, yet faces hurdles in delivering accurate and reliable arrival times. This discrepancy often culminates in delays and a decline in ridership, particularly in areas with a heavy reliance on bus transit. A prevalent challenge is the mismatch between actual bus arrival times and their scheduled counterparts, leading to disruptions in fixed schedules. Our study, utilizing New York City bus data, reveals an average delay of approximately eight minutes between scheduled and actual bus arrival times.
This research introduces an innovative, AI-based, data-driven methodology for predicting bus arrival times at various transit points (stations), offering a collective prediction for all bus lines within large metropolitan areas. Through the deployment of a fully connected neural network, our method elevates the accuracy and efficiency of public bus transit systems. Our comprehensive evaluation encompasses over 200 bus lines and 2 million data points, showcasing an error margin of under 40 seconds for arrival time estimates. Additionally, the inference time for each data point in the validation set is recorded at below 0.006 ms, demonstrating the potential of our Neural-Net based approach in substantially enhancing the punctuality and reliability of bus transit systems.}

\onecolumn \maketitle \normalsize \setcounter{footnote}{0} \vfill

\section{\uppercase{Introduction}}
\label{sec:introduction}

Over the last half-century in the US, the share of workers commuting via public transportation has dwindled. This decline is largely ascribed to governmental separation of land-use development planning from transportation, fueling suburban sprawl, uneven public service distribution, and escalating car reliance in many American cities \cite{freemark2021us,pulugurtha2022does}. Despite concerted efforts over recent decades to bolster public transportation, ridership in the United States hasn’t witnessed a significant uptick, falling below anticipated levels. This stagnation is driven by factors such as urban sprawl, suburbanization, private car ownership, low fuel prices, cutbacks in transit services, and the emergence of ride-hailing giants like Uber and Lyft \cite{erhardt2022has,graehler2019understanding}.

\begin{figure}[ht]
    \centering
    \includegraphics[width=\linewidth]{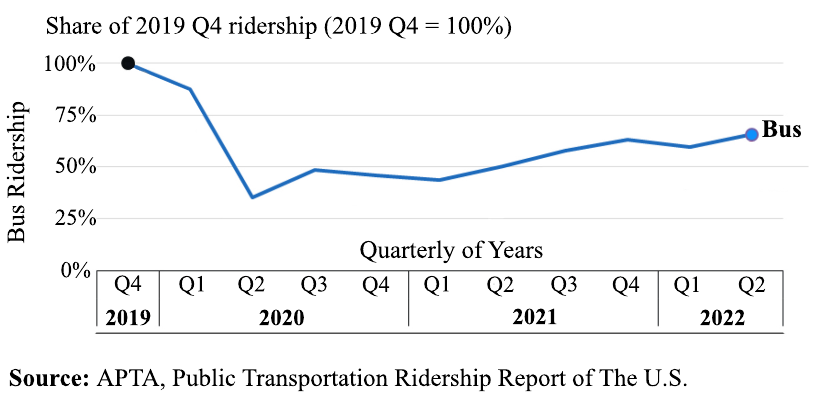}
    \caption{Quarterly Public Bus Transportation Ridership in the U.S. In 2020 and 2021, public transportation ridership was less than half its pre-pandemic level. While bus ridership has recovered somewhat, it was much lower in the second quarter of 2022 than in the final pre-pandemic quarter. Bus ridership for commuters grew by 66\% in the second quarter of 2022 \cite{mallett2022public}.}
    \label{fig1}
\end{figure}

The recent COVID-19 pandemic has further exacerbated the decline in bus ridership across the country, deteriorating the situation from its prior state. The American Public Transportation Association (APTA) notes a stark reduction in transit usage due to the pandemic, with a downturn of over 50 percent between 2019 and 2020 (Figure \ref{fig1}). However, there's a silver lining post-pandemic; public transportation ridership in the U.S. has rebounded from 19\% in April 2020 to 72\% in September 2022, marking the highest level since 2019 \cite{mallett2022public,APTA}.

\begin{figure*}[!ht]
    \centering
    \includegraphics[width=0.8\linewidth]{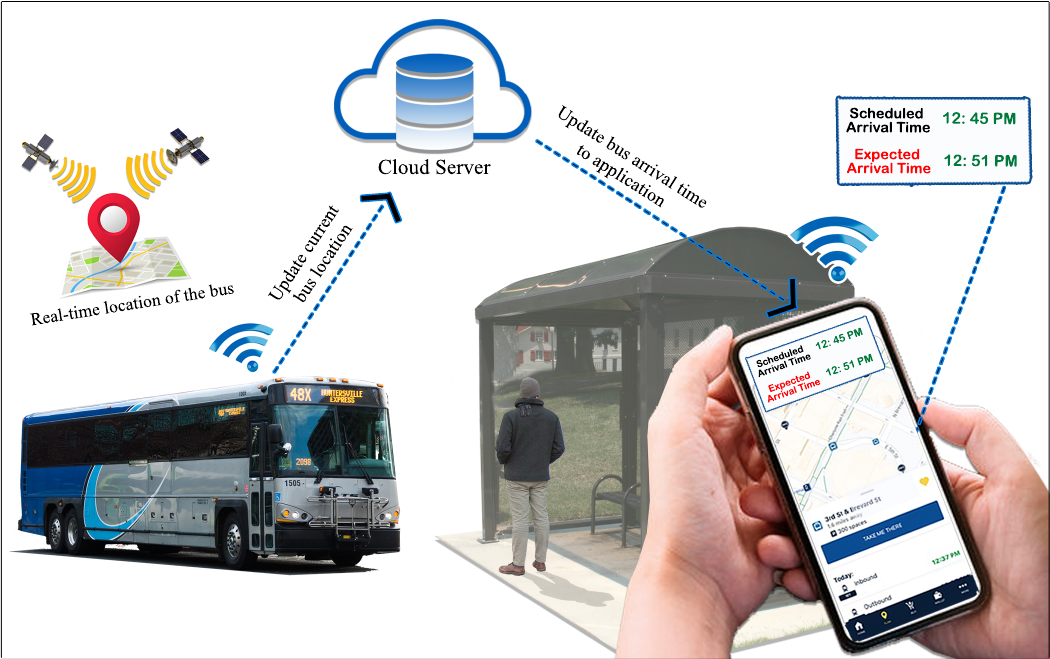}
    \caption{Integrating an AI prediction model into a mobile bus app enhances user experience and operational efficiency. Our model predicts bus arrival times using diverse data sources, providing real-time precision. Users can easily access these predictions via cloud-based services for a reliable travel experience.}
    \label{fig2}
\end{figure*}

The burgeoning discourse around smart cities has piqued the interest of scholars across diverse fields \cite{pazho2023ancilia,gholami2023federated,noghre2022pishgu}. Central to the smart city paradigm is transit reliability, a critical consideration for commuters aiming to minimize long commutes and waiting times on public transit.

To cater to individuals heavily reliant on public transit, developed cities globally are honing their transit scheduling systems. As previously discussed, numerous factors can influence transit ridership, with service predictability being paramount to mitigate undue wait times and enhance trip planning reliability \cite{pulugurtha2022does,sen2022transit}. Unreliable transit services can thwart commuters' travel plans, potentially prompting a shift to alternative transportation modes like personal vehicles. Operational uncertainties and delays may erode transit users' confidence, resulting in reduced ridership and increased dependence on alternative transportation modes. Such unreliable services compel transit users to allocate more time for waiting, culminating in extended wait durations at transit stops \cite{xu2017bus,huang2022bus,zhong2020bus}.

Many cities have rolled out dedicated mobile applications for bus transit, furnishing schedules and aiding passengers in pre-planning their journeys \cite{fu2014bus}. However, the absence of real-time information in these apps often vexes passengers attempting to plan their commute. A sizeable number of commuters resort to other applications (e.g., Google Maps or Waze \cite{splend}) for planning their transit. Nonetheless, these applications, reliant on crowd-sourced information, often fall short in providing requisite accuracy and don't liaise with local bus transits to bolster scheduling and operational efficiency \cite{splend}.

Public transit holds the promise of delivering real-time estimated arrival times akin to private sector ride-sharing platforms like Uber or Lyft \cite{chen2015multi,TSG2019}. Smart, data-driven stratagems could potentially elevate predictability and efficiency across public bus transit systems, mirroring the reliability and predictability seen in Uber/Lyft. The requisite steps encompass regional data gathering, analysis, pattern discernment, and forecasting to augment arrival time accuracy and bus trip planning across the entire network. By doing so, transit authorities could potentially boost ridership, fostering a more sustainable and cost-effective alternative to personal vehicle use, thereby contributing to a greener and more equitable society \cite{diab2015bus,teliacompany}.

This manuscript unveils a deep-learning-centric model for predicting bus arrival times, employing a unified Fully Connected Neural Networks (FCNNs) framework based on historical and environmental data. This model exhibits robust scalability and generalization across numerous bus lines within identical transit networks, surpassing the capabilities of classical machine learning methodologies. Utilizing the New York City Bus System Dataset, encompassing over 200 bus lines and 2 million data points, we conducted our analysis.

Our findings elucidate that our AI-powered model, anchored in FCNNs, significantly curtails the average estimated error in bus arrival times to 40 seconds, a noteworthy improvement compared to the average delay times in the dataset. Per our study, FCNNs outshine traditional machine learning approaches like SVR in tackling transportation conundrums with extensive input features.

The quintessence of this endeavor is to seamlessly assimilate the developed model into existing public bus transit mobile applications, as depicted in Figure \ref{fig2}, with the overarching aim of enriching the bus transit experience. By leveraging this methodology, we aspire to substantially diminish waiting times for passengers, thereby enhancing their commuting experience.

In summary, the contributions of this paper encompass:

\begin{itemize}
\item The introduction of a unified, deep-learning-based Fully Connected Neural Network (FCNN) framework aimed at predicting bus arrival times across numerous bus lines within a singular bus transit network.
\item The thorough assessment of the proposed model's accuracy utilizing the expansive New York City Bus System Dataset, which comprises more than 200 bus lines and 2 million data points.
\item The illustration of our methodology's superior scalability and generalization capabilities when juxtaposed with classical machine learning models such as Support Vector Machines.
\end{itemize}

The rest of this paper is structured as follows: The subsequent section is devoted to bus arrival time prediction literature review. The preliminaries and dataset section contains a detailed description of the dataset used in this study and some exploratory data analysis. Our methodology is illustrated in the proposed neural network methodology section. Finally, our model is validated, and we present our conclusions in the last section.

\section{\uppercase{Related Works}}

Our objective in this section is to assess the efficiency of examples similar to our research demonstrating how data-driven approaches can be used for bus transit systems, arrival time prediction, and scheduling optimization. Public transportation is a crucial component of a connected and smart community. Therefore, citizens demand real-time information regarding transportation assets' arrival and departure. In many cities worldwide, intelligent transportation systems with demand-responsive services are being implemented to bridge the gap between public transportation and private cars. In some early research, data analytics has been used to optimize public bus schedules and minimize passenger wait times.

Different technologies can be utilized that could generate real-time data for bus arrival time prediction. Among them, Global Positioning Systems (GPS), Automatic Passengers Counter Systems (APCS), and Crowdsensing solutions in which users cooperate with the system through a mobile application are the most popular ones \cite{gaikwad2019performance,yin2017prediction}. 

The problem of bus arrival time prediction was studied by considering different models and various essential factors. In a study by N. Gaikwad and S. Varma \cite{gaikwad2019performance}, the crucial features for bus arrival time prediction and standard evaluation metrics were presented. The main factors affecting bus arrival time are the source, destination, bus location coordinates, traffic density, stop-to-stop distance, workday, and so on. 

In another study by Rafidah Md.\cite{noor2020predict}, bus arrival time was predicted using the Support Vector Regression (SVR) model. Petaling Jaya City Bus data was used in this study, including a sequence of bus stations, bus station names, the coordinates of the bus stations, timestamps, and the distance covered from the previous station to the next station. They also implemented their prediction model with and without weather data and showed that adding weather parameters for their dataset shows a negligible difference in their prediction error.

Also, a study by F. Sun, Y. Pan, J. White, and A. Dubey \cite{sun2016real} introduced a public transportation decision support system for short-term and long-term prediction of arrival bus times. This study used the real-world historical data of two Nashville bus system routes. The approach of this research combined the clustering analysis and Kalman filters with a shared route segment model to produce more accurate arrival time predictions and, based on their results, compared to the basic arrival time prediction model that Nashville MTA was using, their system reduced arrival time prediction errors by 25\% on average when predicting the arrival delay an hour ahead and 47\% when predicting within a 15-minute future time window \cite{sun2016real}.

S. Basak, F. Sun, S. Sengupta, and A. Dubey have conducted a similar study \cite{dubey2019data}, using unsupervised clustering mechanisms to optimize transit on-time performance. As a local case study, they analyzed the monthly and seasonal delays of the Nashville metro region and clustered months with similar patterns. In this paper, they presented a stochastic optimization toolchain along with sensitivity analyses for choosing the optimal hyperparameters, and they solved the optimization problem by using a single-objective optimization task as well as a greedy algorithm, a genetic algorithm (GA), and a particle swarm optimization (PSO) algorithm \cite{dubey2019data}.

According to the newest research in \cite{sun2019transit}, dynamic data-driven application systems (DDDAS) that use real-time sensors and a data-driven decision support system can provide online model learning and multi-time-scale analytics to enhance the system's intelligence. As part of their study, the authors analyzed an online bus arrival prediction system in Nashville using historical and real-time streaming data, which can be packaged as modular, distributed, and resilient micro-services. The long-term delay analysis service excludes noise from outliers in historical data to identify delay patterns associated with different hours, days, and seasons for specific time points and route segments. City planners can use the feedback data generated by these analytics services to improve bus schedules and increase rider satisfaction \cite{sun2019transit}. 

In addition, another study by S. Nannapaneni and A. Dubey \cite{nannapaneni2019towards} researched rerouting a single bus to serve spatially and temporally better changing travel demands. The aim was to propose a flexible framework for public transit rerouting. The study was demonstrated on Route 7 of the Nashville Metropolitan Transit Authority (MTA). The authors identified several flex stops with high travel demand using clustering since people living far from bus routes tend to choose alternate transit modes, leading to increased traffic congestion. They categorized the bus stops along the static routes into critical and non-critical stops and added slack time to account for travel delays during the existing static scheduling process. As a result, flexible routes resulted in less additional travel time than available slack time. The effectiveness of rerouting was analyzed using the percentage increase in travel demand \cite{nannapaneni2019towards}.

\section{\uppercase{Preliminaries and Dataset}}

\subsection{Dataset Description}\label{AA}

 The dataset is a critical component of every AI-based system. This study utilizes New York City Bus data \cite{NYC_MTA}. A total of 232 bus lines were inspected to collect this data, and these records were captured in 10-minute increments from 4468 buses.
 
 This dataset was selected due to its rich properties. More than 6 million data generated in a month are included in this dataset. Not only does this data set have a vast number of records, but it also consists of the most relevant parameters to the problem of arrival time prediction. 
 Each record contains the information in the format of 17 fields, including Vehicle location. Longitude, VehicleLocation.Latitude, DestinationLong, DestinationLat, OriginLong, OriginLat, RecordedAtTime, ArrivalTime, ScheduledArrivalTime, DistanceFromStop, OriginName, DestinationName, PublishedLineName, NextStopPointName, ArrivalProximityText, VehicleRef and DirectionRef. The first 6 fields are the current bus location, destination, and origin coordinates. Other field descriptions are as follows: 
\begin{itemize}
    \item RecordedAtTime is the checkpoint time in which the current location of the bus is recorded and used as the bus observation time in this study.
    \item ArrivalTime is the time when the bus arrives at the next stop.
    \item ScheduledArrivalTime is from the published bus timetable, indicating the scheduled time for the bus to arrive at the next stop. 
    \item DistanceFromStop is the distance of the bus from the next stop at the observation time. 
    \item Origin and destination are defined by OriginName and DestinationName.
    \item PublishedLineName represents in which line bus operates. 
    \item NextStopPointName is the name of the next bus stop. 
    \item ArrivalProximityText shows the current status of the bus in terms of a text, including at stop, approaching, and how many miles the bus is away.
    \item VehicleRef is the reference number for every bus whose location is being tracked. 
    \item DirectionRef field indicates inbound or outbound bus direction.
\end{itemize}

\subsection{Cleaning the data and Preprocessing}\label{AA}
 Data is first cleaned and preprocessed to get meaningful concepts from this dataset. Then, the most related features are created, which will be explained in the methodology section. While around 6 million data instances are available in this dataset, they can not be considered logical observations. Since the goal is to predict the arrival time of the bus to the next station, we are only interested in the data points in which the bus is moving between bus stations. So, data is first filtered based on the "ArrivalProximityText" field, and data samples with at-stop value dropped from the data points, where actual arrival time almost equals the bus observation time. By doing the previous steps, around 2 million records were available to work with.

\subsection{Analysis of the Data}\label{AA}
Different indicators can measure the quality of service in public transit infrastructures. On-time performance at stops is an essential factor. Difference time between scheduled and actual bus arrivals has been selected as the top reason people avoid bus transit systems in many cities \cite{dubey2019data}.

\begin{figure}[t]
    \centering
    \includegraphics[width=8.25cm]{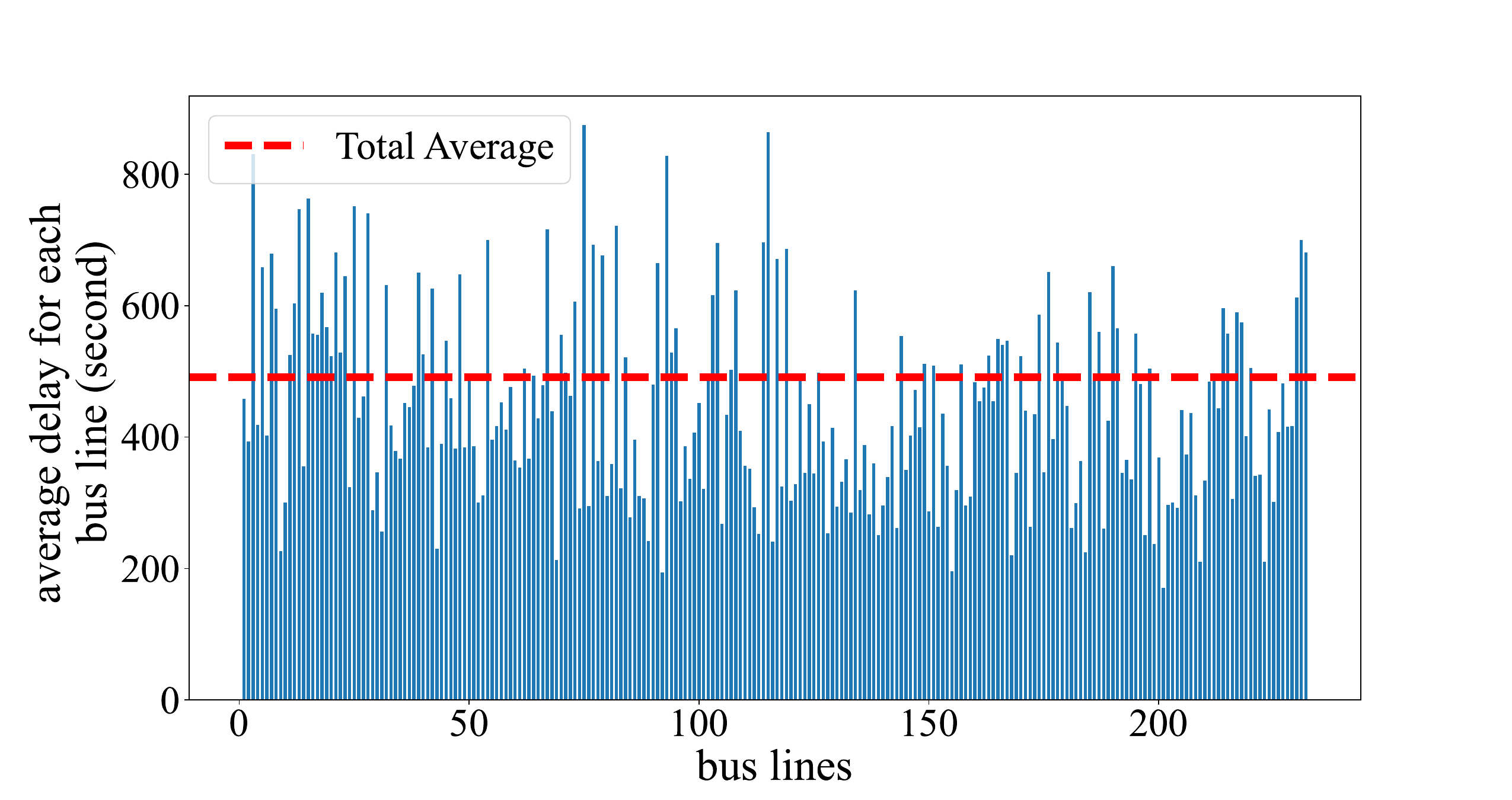}
    \caption{Average delay among all bus lines. Initial analysis of records in the New York dataset shows that the average delay across all bus lines equals 491 seconds.  }
    \label{de_2}
\end{figure}

So, data is first analyzed regarding mismatching between the scheduled arrival time and the actual arrival time. Mismatch time refers to any difference between the bus's scheduled time and arrival time. When the bus arrives at the bus stop earlier, passengers might miss the bus, and also, for late buses, public transportation infrastructure suffers from the delay. Any of these two arrival time variations impact commuters’ satisfaction significantly \cite{dubey2019data}. Our study found that the average delay and mismatch time across all lines of this dataset is around 8 minutes (491 seconds) and 6 minutes, respectively. The average delay for these 232 bus lines has been illustrated in Figure \ref{de_2}.

\section{\uppercase{Proposed Neural Network Methodology}}
\subsection{Feature Extraction}\label{AA}
The New York data set has 232 lines, and each line has been segmented into the number of bus stops. Assigning each line an integer value would not be a practical approach since an ordered relationship exists between integer values and may lead to poor performance of the model. One-hot encoding applies to categorical variables like bus lines without an ordinal relationship. This encoding helps the bus lines be injected into the model in terms of binary variables. Applying one-hot encoding on bus lines expands the input features and adds 232 more inputs. On the other hand, bus stops have some order, and they are fed into the model through integer encoding. The bus stop input variable can help the model track the traffic conditions and passenger flow varying from one bus stop to another.

As mentioned before, the bus records in this dataset were collected for a month. Because of the wide time variation, time is injected into the model in terms of two categories rather than feeding it directly into the model. This approach avoids injecting a lot of noise into the model. 

First, based on the day of the bus operation, the variable "day type" is added to the input features, which can get two values, "weekend" and "workday". The other time-related variable is the rush hour status. According to the operation time of the bus, this feature assigns to each record of the dataset, determining whether the bus operates during rush hour or not. Rush hour in New York spans from 6 AM to 10 AM and 3 PM to 7 PM \cite{MTA}.  

In addition to the features that were previously mentioned, there are two distance-related features in the model. The distance input feature, the most important feature among other features, indicates how many meters the bus is far from the next stop. Far status is another distance-related feature added to input features according to the distance value. It is a binary feature that changes depending on whether the distance is below or above a specified threshold. Research on the distribution of bus stop spacings in the United States reveals that the average distance between bus stops in New York is 328 meters \cite{pandey2021distributions}. In our study, we determined a threshold value of 250 meters through trial and error. when the distance is below this threshold, it indicates that the bus is on its way to the next station and probably not waiting at the previous bus stop. 

Trip time ($Tr$) is the target variable we aim to predict with our model, representing the time required for a bus to reach the next bus station from its current location. It is calculated in seconds by subtracting the bus observation time ($T_{ob}$), which corresponds to the RecordedAtTime in the dataset, from the actual arrival time ($T_{ar}$), associated with ArrivalTime in the dataset. In practical scenarios, knowing the trip time allows for the calculation of the arrival time of the bus.
	\begin{equation} \label{eqn}
	Tr = {T_{ar}} - {T_{ob}}
	\end{equation}  
Table~\ref{tab_NN_3} summarizes the input and output features that were produced during the feature extraction step.

\begin{table}[!ht]
\begin{center}
\caption{Input and Output Features}
\begin{tabular}{|l|c|}
\hline
Input Features & \begin{tabular}[c]{@{}c@{}}One-hot Encoded Bus Lines \\ Distance \\ Day Type \\ Rush Hour Status \\ Bus Stops \\ Far Status\end{tabular} \\ \hline
Output Feature & Trip Time                                                      \\ \hline
\end{tabular}
\label{tab_NN_3}
\end{center}
\end{table}

\subsection{Feature Scaling}\label{AA}
Due to the different range of input features, data needs to be scaled prior to being injected into the model. Some input features, like rush hour status, are in the binary form and represented by 1 and 0, while others like distance, can be hundreds of meters. Without feature scaling, the model can be affected by the different range of features, assigning higher weights to the features with large scale. So, Min-Max scaling is used to transform the value of all input features to the range of 0 and 1.  
\subsection{Train and Validation Sets}\label{AA}
The dataset is divided into a train and validation set. 80\% of the dataset has been categorized as a training set for the training of our model, while 20\% of the dataset has been used as a validation set. The total number of instances is 2.13 million. 1.7 million is used for training our model, while the rest is utilized for validation. It is also worth mentioning that the average of training data samples for each line is 7327. 

\subsection{Model Design}\label{AA}
Artificial Neural Networks (ANNs) are very common in forecasting bus trip time. Previous studies have demonstrated that ANNs are effective in predicting nonlinear relationships in complicated problems. \cite{bai2015dynamic}.

In this study, we make use of FCNNs to predict the bus trip time. As discussed in the previous section, due to the large number of bus lines, FCNNs can handle high-dimensional feature spaces by using hidden layers and non-linear activation functions. To determine the optimal architecture for our model, we conducted various experiments with different configurations, including the number of hidden layers, neurons in each layer, and activation functions. Based on the results obtained, we selected the best-performing model with enhanced prediction capabilities. 

\begin{figure}[htp]
    \centering
    \includegraphics[width=\linewidth]{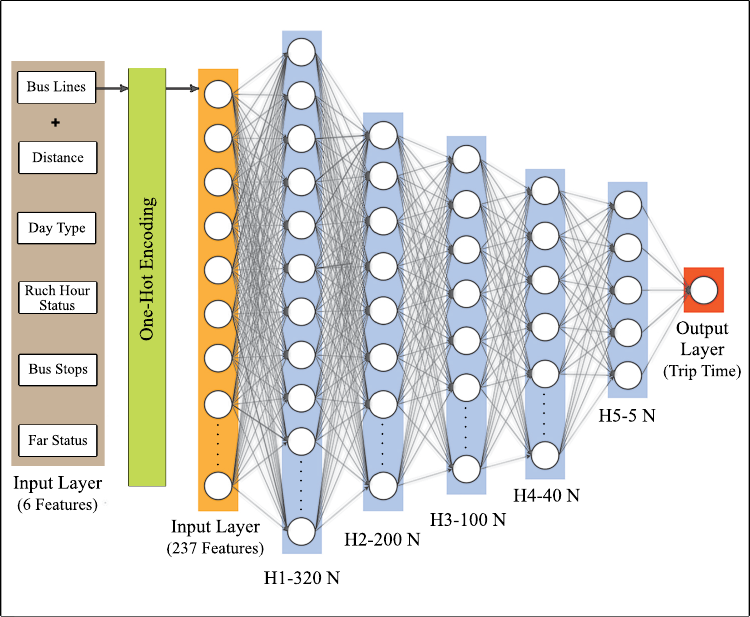}
    \caption{Structure of our model based on the Fully Connected Neural Network. One-hot encoding applies to bus lines and extends it to 232 features. These converted features with other 5 features, including distance, day type, rush hour status, bus stops, and far status feed to the Fully Connected Neural Network. The proposed model consists of 5 hidden layers and ReLU as an activation function. The number of neurons in each hidden layer can also be seen in the figure. H1-320N indicates that the first hidden layer consists of 320 neurons. }
    \label{NN Model-2}
\end{figure}

As illustrated in Figure \ref{NN Model-2}, the model is fed with 237 input features, including 232 features generated through one-hot encoding for bus lines, distance, day type, rush hour status, bus stops, and far status. Additionally, the output layer consists of one neuron for predicting the bus trip time based on transformed input data from the preceding hidden layers. Consequently, throughout all our experiments, the input layer and output layer remained identical.

Our experimental methodology involved a thorough investigation into the architecture of the neural network. Specifically, we systematically varied the number of hidden layers from 2 to 7, evaluating their impact on model performance. Within each configuration, we adjusted the number of neurons in a descending order across layers, with the first layer having the maximum number of neurons and the last layer having the minimum number of neurons. By modifying these architectural parameters, our goal was to create a balance between model complexity and generalization. This aimed to ensure that the FCNN captures intricate patterns in the data without overfitting.

Our experiments demonstrated that surpassing 5 hidden layers fails to enhance accuracy. Consequently, we utilized 5 hidden layers for our model. Concerning the number of neurons for each layer, we observed that an increased number of neurons, 512 neurons, did not yield an improvement in accuracy. Instead, it led to a more complex model with more parameters without any benefit in predictive performance. Consequently, we settled on 320 neurons for the first layer, ensuring an optimal balance between capturing complexity and preventing unnecessary parameter inflation. The same rationale guided our decisions in determining the most suitable number of neurons for each layer.

Moreover, another important factor in FCNNs is the choice of activation function that plays a key role by introducing non-linearity to the model. Rectified Linear Unit (ReLU) function is used as an activation function in our model. Since the number of hidden layers in our model is large, ReLU is a better choice than Sigmoid and Hyperbolic Tangent (Tanh), helping to mitigate the vanishing gradient problem.

According to Figure \ref{NN Model-2}, our best model has seven layers, including an input layer, 5 hidden layers, and an output layer.  In the first hidden layer, the model learns more complex representations of input features by increasing neurons to 320. The number of neurons decreases step by step in the next hidden layers, from 200 in the second hidden layer to 5 neurons in the fifth hidden layer. The structure of the presented model can be seen in Table~\ref{tab_NN_1}.

\begin{table}[htpb]
\centering
\caption{Structure of fully connected neural network applied to New York dataset with 232 bus lines }
\begin{tabular}{|cc}
\hline
\multicolumn{2}{c}{\textbf{Parameters}} \\
\hline \hline
\multicolumn{1}{c|}{Number of inputs} & 237\\ \hline
\multicolumn{1}{c|}{Number of hidden layers} & 5 \\ \hline
\multicolumn{1}{c|}{ Activation function}& ReLU \\ \hline
\multicolumn{1}{c|}{Number of neurons in the first layer }& 320 \\ \hline
\multicolumn{1}{c|}{Number of neurons in the second layer }& 200 \\ \hline
\multicolumn{1}{c|}{Number of neurons in the third layer}& 100\\ \hline
\multicolumn{1}{c|}{Number of neurons in the fourth layer }& 40\\ \hline
\multicolumn{1}{c|}{Number of neurons in the fifth layer}& 5\\ \hline
\multicolumn{1}{c|}{Number of outputs} & 1\\ \hline
\end{tabular}
\label{tab_NN_1}
\end{table}
\section{\uppercase{Results and Discussion}}
 \subsection{Performance Measurements}\label{AA}
The performance evaluation of arrival time predicted by the model can be done using different measures, including Mean Absolute Percentage Error (MAPE), Mean Square Error (MSE), and Root Mean Square Error (RMSE). \newline
In this study, we assess the model's accuracy using RMSE, which quantifies the difference between predicted trip times and actual trip times in seconds. RMSE is a widely used metric in the field of bus arrival time analysis, facilitating comparisons with other models. Furthermore, RMSE shares the same unit as the predicted values (seconds), simplifying the interpretation of errors in terms of time. 

RMSE can be represented as the following equation where $t_{act}$ is the actual bus trip time, (Tr in Equation \ref{eqn}), $t_{pred}$ stands for the predicted bus trip time based on the proposed model, and n is the sample size for prediction. Lower RMSE represents better performance in prediction. 

\begin{equation} \label{eqn1}
	RMSE=\sqrt \frac {\Sigma^{n}_{i=1}  (t_{act} - t_{pred})^2} {n}
	\end{equation}

\subsection{Results and Model Performance Discussion}\label{AA}

 \subsubsection{Results for Fully connected NN on all bus lines}\label{AA} 
 The training process was implemented on a system with an Intel i7-1185G7 processor with 4 cores and a speed of 3.00 GHz.
 Table~\ref{tab_NN_0} shows the results of applying our model on 232 bus lines with the learning rate of $1e-2$.

\begin{table}[htpb]
\centering
\caption{Obtained results in terms of RMSE for the New York dataset with 232 bus lines}
\begin{tabular}{|cc}
\hline
\multicolumn{2}{c}{\textbf{Results}} \\
\hline \hline
\multicolumn{1}{c|}{Training RMSE} & 35.69 s\\ \hline
\multicolumn{1}{c|}{Validation RMSE} & 35.74 s \\ \hline
\end{tabular}
\label{tab_NN_0}
\end{table}

\begin{figure}[t]
    \centering
    \vspace{-15pt}
    \includegraphics[width=8.25cm]{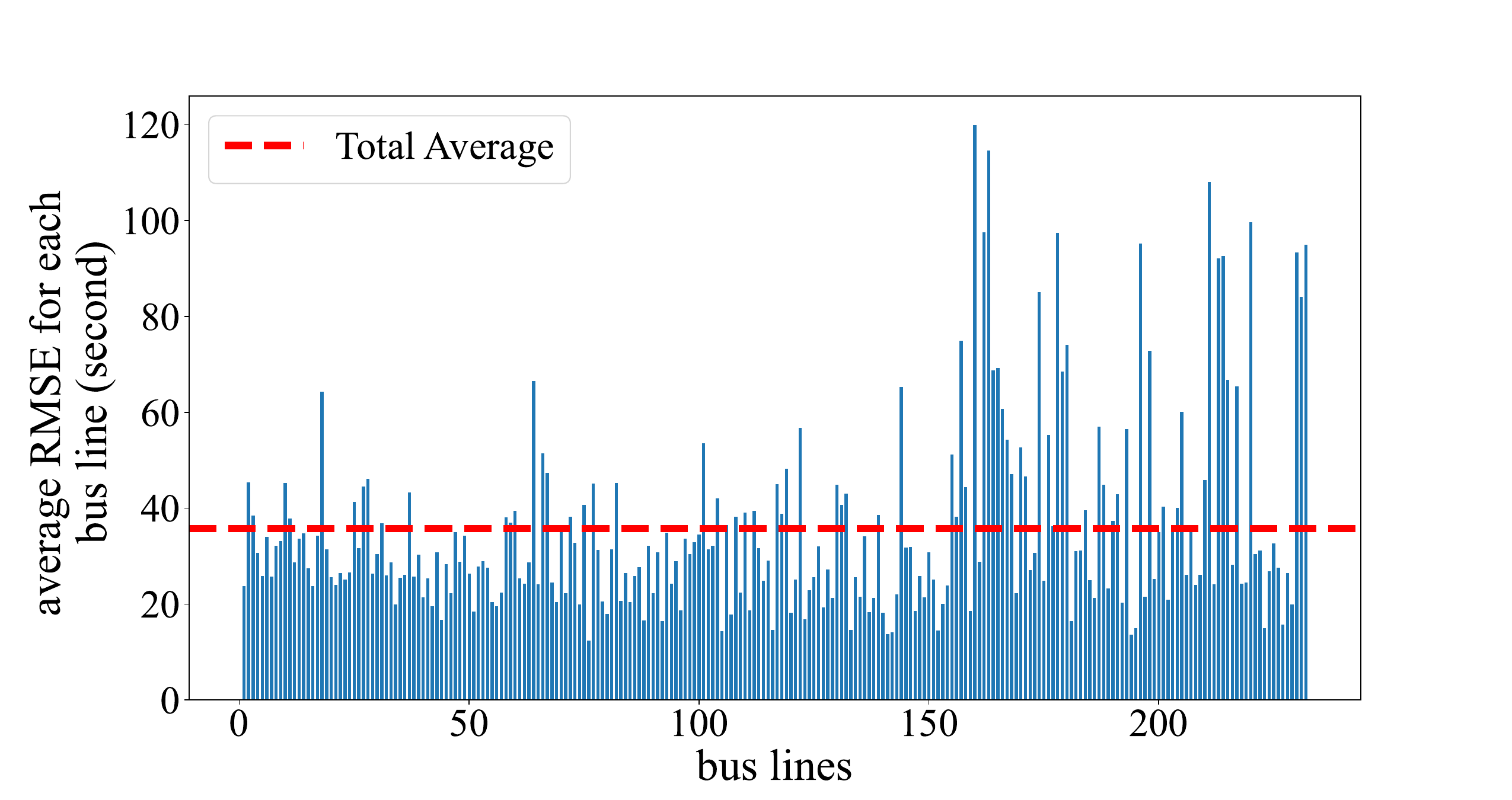}
    \caption{Performance of the model for all bus lines. The average RMSE across all bus lines in the validation set is 35.74 seconds. Among the prediction error values, bus line 160 has the greatest RMSE, with a value of 119.99 seconds. In contrast, the lowest error belongs to bus line 76 with an RMSE equal to 12.42 seconds.}
    \label{fig:RMSE_validation}
\end{figure}

 It can be observed that the average RMSE for all bus lines is 35.74 seconds. In other words, the predicted arrival time of the bus to the next station has an error lower than 36 seconds. This prediction error can be contrasted with the average delay observed across all bus lines in the dataset, which equals to 491 seconds according to the data analysis section. Figure \ref{fig:RMSE_validation} demonstrates the RMSE over each bus line. While the highest prediction error is 119.99 seconds in line number 160, the lowest RMSE belongs to line number 76, with the RMSE equal to 12.42 seconds. 
 
\begin{figure}[htpb]
    \centering
    \includegraphics[width=7.4cm]{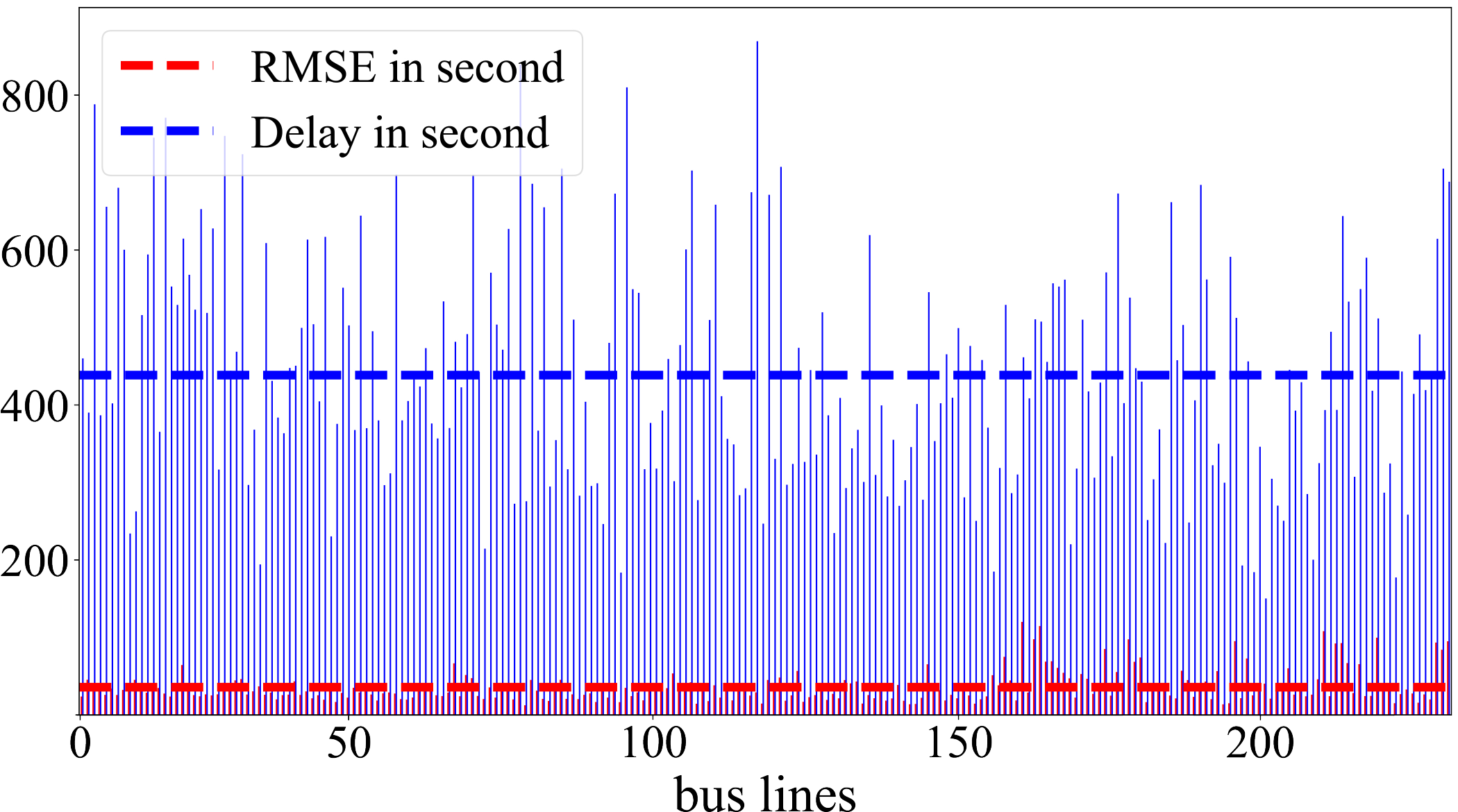}
    \caption{Comparison between actual delay and RMSE across validation set samples, containing 426,323 samples. The average delay for all bus lines in the validation set is 438 seconds, while the prediction error over these samples is less than 36 seconds.}
    \label{fig:RMSE_validation1}
\end{figure}

Additionally, Figure \ref {fig:RMSE_validation1} illustrates the comparison between the actual delay and RMSE of the predicted arrival time across all bus lines in the validation set. The large RMSE values in certain bus lines compared to others could be due to the lack of relevant features in predicting bus trip time. There is a wide range of other factors affecting the bus trip time, but not available in this dataset. For instance, passenger demand is a feature that this dataset does not include. By equipping buses with passenger counting systems, passenger demand for each bus stop can also be recorded. This parameter impacts the bus dwell time, referring to a bus's time at a stop without moving. Additionally, a potential area for future research could involve investigating how weather types can influence error arrival time prediction.

\begin{table}[htpb]
\centering
\caption{Properties of the proposed model for bus arrival time prediction on the New York dataset}
\begin{tabular}{|cc}
\hline
\multicolumn{2}{c}{\textbf{Model Properties}} \\
\hline \hline
\multicolumn{1}{c|}{Total Training Time} & 7171s\\ \hline
\multicolumn{1}{c|}{Total Inference Time} & 2.42s\\ \hline
\multicolumn{1}{c|}{\makecell{Inference Time per each \\ Validation Sample}} & 0.00578 ms\\ \hline 
\multicolumn{1}{c|}{Number of Parameters} & 164710\\ \hline
\multicolumn{1}{c|}{Computational Complexity } &  165380 Mac \\ \hline
\end{tabular}
\label{tab_NN_4}
\end{table}

\begin{figure}[htpb]
    \centering
    \includegraphics[width=7.4cm]{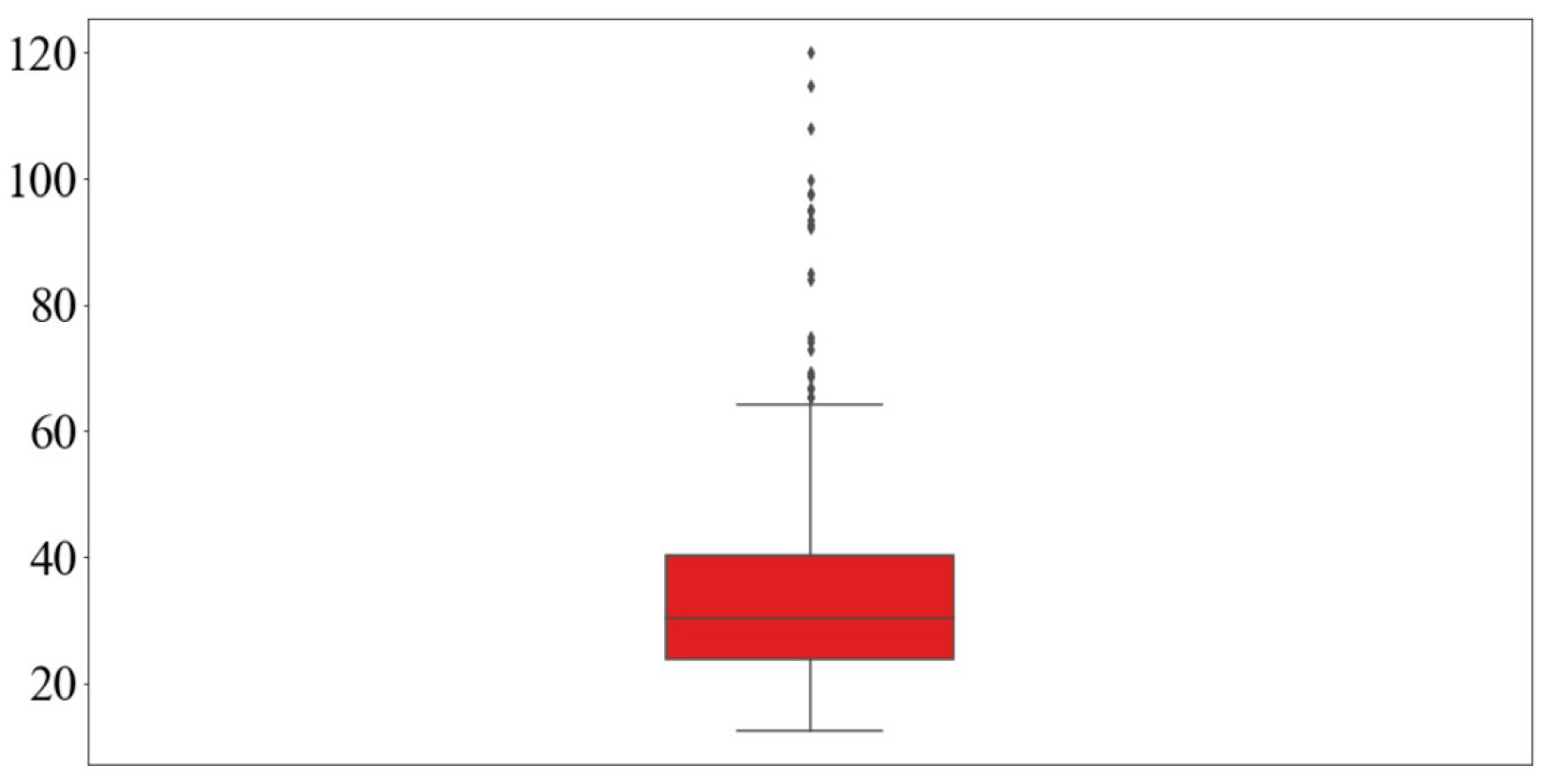}
    \caption{RMSE distribution in the form of a boxplot. It displays the difference in RMSE values by showing the median of 35.74 seconds.}
    \label{fig:box}
\end{figure}

In Figure \ref{fig:box}, we have also shown RMSE distribution. Training time and inference time per each validation set data point are also presented in Table~\ref{tab_NN_4}. The average inference time for each validation data point is 0.00578 ms. This implies if a passenger sends a request to the cloud to get the bus arrival time for their trip, it takes less than 0.006 ms to produce AI-based predictions. It should be noted this inference time indicates the required time only for one access. When thousands of passengers request bus arrival time to the cloud, it will grow significantly.

\subsubsection{Scalability Comparison between our model and SVR }\label{AA}

Since there are more than 200 lines in the dataset, a generalized model is needed to predict the arrival time with the lowest possible error for all bus lines.
This section illustrates the scalability comparison of our model and SVR for this dataset. The reason for making this comparison is that SVR is among the other machine learning approaches that are popular for bus arrival time prediction problems, and a lot of researchers used SVR with the Radial Basis Function (RBF) kernel for bus arrival time prediction \cite{noor2020predict}.  

\begin{figure}[t]
    \centering
    \vspace{-15pt}
    \includegraphics[width=8.25cm]{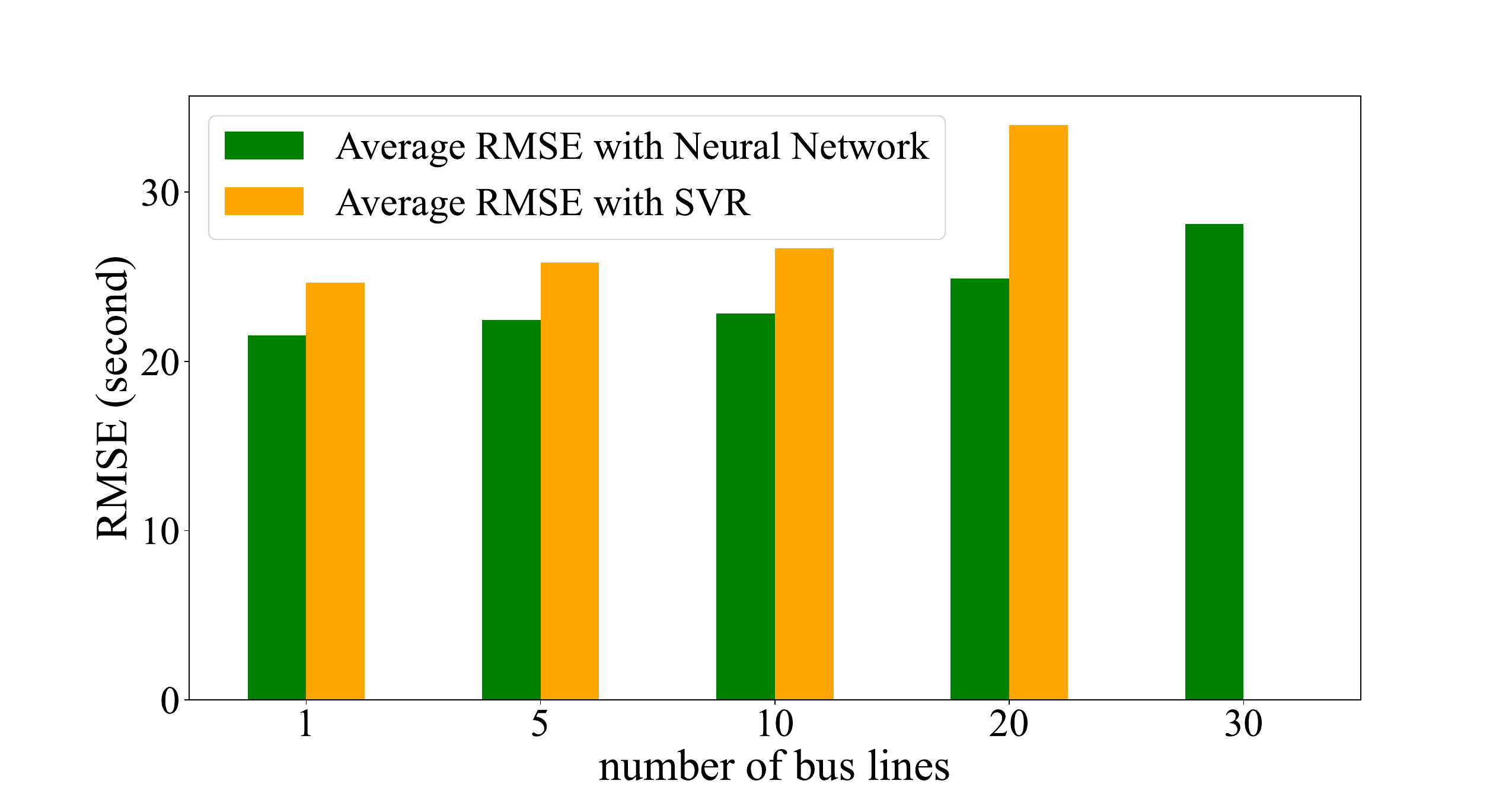}
    \caption{Scalability comparison of FCNN and SVR. These two models were evaluated for different numbers of bus lines. In the range of 1 to 20 bus lines, the accuracy of SVR prediction decreased significantly, while NN performance remained almost unchanged. When the number of bus lines exceeds 30, SVR can not be trained on this dataset.}
    \label{SVR_NN_2}
\end{figure}

So, for a different number of bus lines, SVR with RBF kernel was used. The experimental results, as observed in Figure~\ref{SVR_NN_2}, indicate that in a small number of lines, SVR and our model prediction patterns are almost the same. RMSE for prediction on 10 lines using FCNN and SVR is 22.84 and 26.67, respectively. When the number of lines rises from 10 to 20, RMSE is 24.90 and 33.98, showing a notable increase in RMSE for the SVR model, and when the number of bus lines surpasses 30, SVR becomes untrainable on this dataset. 
Hence, in terms of scalability, our model has a better prediction ability than SVR, which is why FCNN was selected for the whole dataset. 
 
\section{\uppercase{Conclusions and Future Work}}
In this study, we engineered an AI-driven prediction model aimed at propelling bus transit systems into a realm of enhanced intelligence, thereby significantly elevating passenger experience by curtailing protracted wait times. Our innovative blueprint unfolds a real-time bus arrival prediction mechanism, presenting a stark contrast to the conventional rigidity of fixed schedules.

The predictive model assimilates various input features encompassing bus lines, distance, day type, rush hour status, bus stops, and far status. The culmination of our endeavor, rooted in the Fully Connected Neural Networks (FCNNs) framework, manifested in an average estimated error reduction to less than 40 seconds across all bus lines within the dataset. This outcome heralds a substantial leap forward when juxtaposed against the average delay time embedded in the dataset. Our forthcoming stride is geared towards melding this AI-centric model within a smart mobile application, thereby furnishing real-time insights to commuters on the go.

The scope of this paper was partially tethered to select features pertinent to bus trip time, dictated by the constraints inherent in the utilized dataset. As we move forward, numerous opportunities for future research in this domain beckon exploration. Firstly, investigating the integration of other factors, such as passenger flow, and meteorological conditions, could provide a more comprehensive understanding of the factors influencing bus arrival times. Additionally, delving into alternative architectures, particularly self-attention based neural networks, could enhance the model's adaptability to diverse transportation datasets. The innate capacity of such models to capture long-term temporal dependencies within the bus data suggests their potential for having more accurate and efficient forecasting techniques in transportation systems.

\section{\uppercase{Acknowledge}} 
This research is supported by the UNC Charlotte's College of Engineering seed grant, the UNC Charlotte's Urban Institute, and the UNC Charlotte's School of Data Science.


\bibliographystyle{apalike}
{\small
\bibliography{example}}

\begin{thebibliography}{}

\bibitem[NYC, 2017]{NYC_MTA}
 (2017).
\newblock New york city bus data.

\bibitem[tel, 2017]{teliacompany}
 (2017).
\newblock Smart public transport – key to solving the urban challenge - telia company.
\newblock Accessed on Mar. 12, 2023.

\bibitem[TSG, 2019]{TSG2019}
 (2019).
\newblock Transport, data analytics and ai: Why tfl’s latest initiative is good news.
\newblock Accessed on Mar. 12, 2023.

\bibitem[APT, 2022]{APTA}
 (2022).
\newblock September 2022 apta public transportation ridership update.
\newblock Accessed on Mar. 15, 2023.

\bibitem[spl, 2023]{splend}
 (2023).
\newblock Google maps vs waze – which one is better?
\newblock Accessed on Mar. 15, 2023.

\bibitem[MTA, 2023]{MTA}
 (2023).
\newblock New york city transit key performance metrics.
\newblock Accessed on Mar. 15, 2023.

\bibitem[Bai et~al., 2015]{bai2015dynamic}
Bai, C., Peng, Z.-R., Lu, Q.-C., and Sun, J. (2015).
\newblock Dynamic bus travel time prediction models on road with multiple bus routes.
\newblock {\em Computational intelligence and neuroscience}, 2015:63--63.

\bibitem[Chen et~al., 2015]{chen2015multi}
Chen, C., Chen, W., and Chen, Z. (2015).
\newblock A multi-agent reinforcement learning approach for bus holding control strategies.
\newblock {\em Advances in Transportation Studies}.

\bibitem[Diab et~al., 2015]{diab2015bus}
Diab, E.~I., Badami, M.~G., and El-Geneidy, A.~M. (2015).
\newblock Bus transit service reliability and improvement strategies: Integrating the perspectives of passengers and transit agencies in north america.
\newblock {\em Transport Reviews}, 35(3):292--328.

\bibitem[Dubey et~al., 2019]{dubey2019data}
Dubey, A.~D., Basak, S.~B., Sengupta, S.~S., and Sun, F.~S. (2019).
\newblock Data-driven optimization of public transit schedule.

\bibitem[Erhardt et~al., 2022]{erhardt2022has}
Erhardt, G.~D., Hoque, J.~M., Goyal, V., Berrebi, S., Brakewood, C., and Watkins, K.~E. (2022).
\newblock Why has public transit ridership declined in the united states?
\newblock {\em Transportation research part A: policy and practice}, 161:68--87.

\bibitem[Freemark, 2021]{freemark2021us}
Freemark, Y. (2021).
\newblock Us public transit has struggled to retain riders over the past half century. reversing this trend could advance equity and sustainability. urban institute.
\newblock {\em The Urban Institute}.

\bibitem[Fu et~al., 2014]{fu2014bus}
Fu, J., Wang, L., Pan, M., Zuo, Z., and Yang, Q. (2014).
\newblock Bus arrival time prediction and release: system, database and android application design.
\newblock In {\em International Conference on Algorithms and Architectures for Parallel Processing}, pages 404--416. Springer.

\bibitem[Gaikwad and Varma, 2019]{gaikwad2019performance}
Gaikwad, N. and Varma, S. (2019).
\newblock Performance analysis of bus arrival time prediction using machine learning based ensemble technique.
\newblock In {\em Proceedings 2019: Conference on Technologies for Future Cities (CTFC)}.

\bibitem[Gholami et~al., 2023]{gholami2023federated}
Gholami, S., Lim, J.~I., Leng, T., Ong, S. S.~Y., Thompson, A.~C., and Alam, M.~N. (2023).
\newblock Federated learning for diagnosis of age-related macular degeneration.
\newblock {\em Frontiers in Medicine}, 10.

\bibitem[Graehler et~al., 2019]{graehler2019understanding}
Graehler, M., Mucci, R.~A., and Erhardt, G.~D. (2019).
\newblock Understanding the recent transit ridership decline in major us cities: Service cuts or emerging modes.
\newblock In {\em 98th Annual Meeting of the Transportation Research Board, Washington, DC}.

\bibitem[Huang et~al., 2022]{huang2022bus}
Huang, H., Huang, L., Song, R., Jiao, F., and Ai, T. (2022).
\newblock Bus single-trip time prediction based on ensemble learning.
\newblock {\em Computational intelligence and neuroscience}, 2022.

\bibitem[Mallett, 2022]{mallett2022public}
Mallett, W.~J. (2022).
\newblock Public transportation ridership: Implications of recent trends for federal policy.
\newblock Technical report.

\bibitem[Nannapaneni and Dubey, 2019]{nannapaneni2019towards}
Nannapaneni, S. and Dubey, A. (2019).
\newblock Towards demand-oriented flexible rerouting of public transit under uncertainty.
\newblock In {\em Proceedings of the Fourth Workshop on International Science of Smart City Operations and Platforms Engineering}, pages 35--40.

\bibitem[Noghre et~al., 2022]{noghre2022pishgu}
Noghre, G.~A., Katariya, V., Pazho, A.~D., Neff, C., and Tabkhi, H. (2022).
\newblock Pishgu: Universal path prediction architecture through graph isomorphism and attentive convolution.
\newblock {\em arXiv preprint arXiv:2210.08057}.

\bibitem[Noor et~al., 2020]{noor2020predict}
Noor, R.~M., Yik, N.~S., Kolandaisamy, R., Ahmedy, I., Hossain, M.~A., Yau, K.-L.~A., Shah, W.~M., and Nandy, T. (2020).
\newblock Predict arrival time by using machine learning algorithm to promote utilization of urban smart bus.

\bibitem[Pandey et~al., 2021]{pandey2021distributions}
Pandey, A., Lehe, L., and Monzer, D. (2021).
\newblock Distributions of bus stop spacings in the united states.
\newblock {\em Findings}.

\bibitem[Pazho et~al., 2023]{pazho2023ancilia}
Pazho, A.~D., Neff, C., Noghre, G.~A., Ardabili, B.~R., Yao, S., Baharani, M., and Tabkhi, H. (2023).
\newblock Ancilia: Scalable intelligent video surveillance for the artificial intelligence of things.
\newblock {\em IEEE Internet of Things Journal}.

\bibitem[Pulugurtha et~al., 2022]{pulugurtha2022does}
Pulugurtha, S.~S., Mishra, R., and Jayanthi, S.~L. (2022).
\newblock Does transit service reliability influence ridership?

\bibitem[Sen et~al., 2022]{sen2022transit}
Sen, R., Bharati, A.~K., Khaleghian, S., Ghosal, M., Wilbur, M., Tran, T., Pugliese, P., Sartipi, M., Neema, H., and Dubey, A. (2022).
\newblock E-transit-bench: simulation platform for analyzing electric public transit bus fleet operations.
\newblock In {\em Proceedings of the Thirteenth ACM International Conference on Future Energy Systems (e-Energy 2022)}.

\bibitem[Sun et~al., 2019]{sun2019transit}
Sun, F., Dubey, A., White, J., and Gokhale, A. (2019).
\newblock Transit-hub: A smart public transportation decision support system with multi-timescale analytical services.
\newblock {\em Cluster Computing}, 22:2239--2254.

\bibitem[Sun et~al., 2016]{sun2016real}
Sun, F., Pan, Y., White, J., and Dubey, A. (2016).
\newblock Real-time and predictive analytics for smart public transportation decision support system.
\newblock In {\em 2016 IEEE International Conference on Smart Computing (SMARTCOMP)}, pages 1--8. IEEE.

\bibitem[Xu and Ying, 2017]{xu2017bus}
Xu, H. and Ying, J. (2017).
\newblock Bus arrival time prediction with real-time and historic data.
\newblock {\em Cluster Computing}, 20:3099--3106.

\bibitem[Yin et~al., 2017]{yin2017prediction}
Yin, T., Zhong, G., Zhang, J., He, S., and Ran, B. (2017).
\newblock A prediction model of bus arrival time at stops with multi-routes.
\newblock {\em Transportation research procedia}, 25:4623--4636.

\bibitem[Zhong et~al., 2020]{zhong2020bus}
Zhong, G., Yin, T., Li, L., Zhang, J., Zhang, H., and Ran, B. (2020).
\newblock Bus travel time prediction based on ensemble learning methods.
\newblock {\em IEEE Intelligent Transportation Systems Magazine}, 14(2):174--189.

\end{thebibliography}

\end{document}